\pdfoutput=1
\documentclass[11pt]{article}

\usepackage[preprint]{acl}

\usepackage{times}
\usepackage{latexsym}

\usepackage[T1]{fontenc}
\usepackage[utf8]{inputenc}

\usepackage{microtype}

\usepackage{inconsolata}

\usepackage{graphicx}

\usepackage{multirow}
\usepackage{booktabs}
\usepackage{amsmath}
\usepackage{amssymb}
\usepackage{enumitem}
\usepackage{xcolor}
\usepackage{colortbl}
\usepackage{url}
\definecolor{mycell}{gray}{.95}

\setlist[itemize]{noitemsep,nolistsep}
\title{PROPER: A Progressive Learning Framework for Personalized Large Language Models with Group-Level Adaptation}

\author{Linhai Zhang$^{1}$,
        Jialong Wu$^{2}$,
        Deyu Zhou$^{2}$,
        Yulan He$^{1,3}$ \\
        $^1$King's College London \hspace{0.5cm}
        $^2$Southeast University  \hspace{0.5cm}
        $^3$The Alan Turing Institute \\
        \texttt{\{linhai.zhang, yulan.he\}@kcl.ac.uk} \\
        \texttt{\{jialongwu, d.zhou\}@seu.edu.cn}
  }

\begin{document}
\maketitle
\begin{abstract}
Personalized large language models (LLMs) aim to tailor their outputs to user preferences. 
Recent advances in parameter-efficient fine-tuning (PEFT) methods have highlighted the effectiveness of adapting population-level LLMs to personalized LLMs by fine-tuning user-specific parameters with user history.
However, user data is typically sparse, making it challenging to adapt LLMs to specific user patterns.
To address this challenge, we propose PROgressive PERsonalization (PROPER), a novel progressive learning framework inspired by meso-level theory in social science. 
PROPER bridges population-level and user-level models by grouping users based on preferences and adapting LLMs in stages. 
It combines a Mixture-of-Experts (MoE) structure with Low Ranked Adaptation (LoRA), using a user-aware router to assign users to appropriate groups automatically. 
Additionally, a LoRA-aware router is proposed to facilitate the integration of individual user LoRAs with group-level LoRAs.
Experimental results show that PROPER significantly outperforms SOTA models across multiple tasks, demonstrating the effectiveness of our approach.
\footnote{Our code is available at \url{https://anonymous.4open.science/r/PROPER-63E5/}.}
\end{abstract}

\section{Introduction}

Though large language models (LLMs) have demonstrated superior performance across various tasks~\cite{zhao2023survey,chen2024large}, they primarily offer a ``\textit{one-size-fits-all}'' service, which falls short of adapting to individual user preferences.
Personalized LLMs, aimed at tailoring their outputs to user-specific preferences, have therefore become a hot research topic~\cite{salemi-etal-2024-lamp,mysore-etal-2024-pearl,tan-etal-2024-democratizing}.

\begin{figure}[t]
    \centering
    \includegraphics[width=0.42\textwidth]{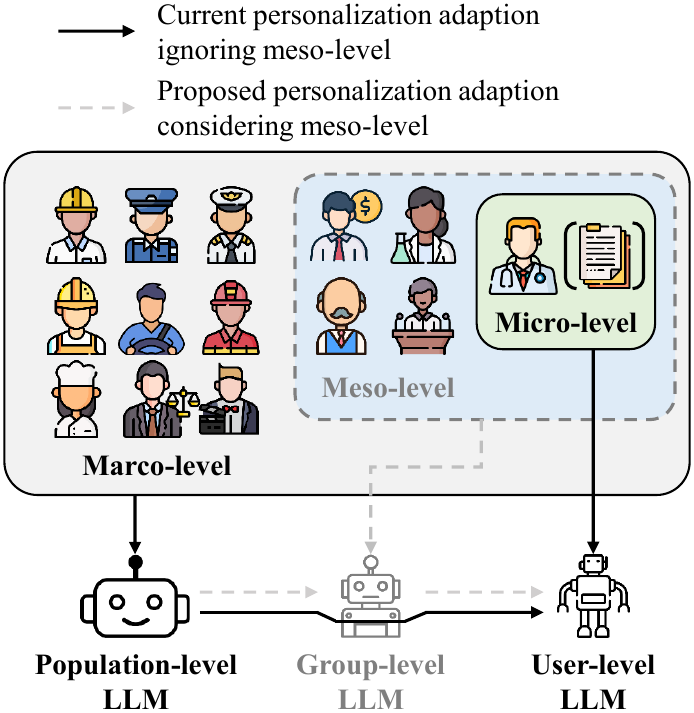}
    \caption{The comparison between different paradigms of LLM personalization, the solid line represents the current paradigms, which adapt the population-level LLM directly to the user-level LLM, while the dashed line illustrates the proposed paradigms, which \textbf{adapt progressively through a group-level LLM} using meso-level data as a bridge.}
    \label{fig:intro}
\end{figure}

Early efforts to personalizing LLMs focused on incorporating user history into prompts using in-context learning~\cite{dai2023uncovering}, retrieval-augmented generation~\cite{mysore-etal-2024-pearl}, and profile-augmented generation~\cite{richardson2023integrating}. 
However, these prompt-based methods struggle with ensuring user data privacy and have limited generalization capabilities~\cite{tan-etal-2024-democratizing}. 
Recent research has shifted towards fine-tuning personalized LLMs, where the base LLMs are fine-tuned on user history to better capture individual preferences.
\citet{tan-etal-2024-democratizing} first proposed to store user-specific preferences and behavior patterns in personalized Parameter-Efficient Fine-Tuning (PEFT) parameters (\textit{e.g.}, LoRA~\cite{hu2022lora}) to enable computationally efficient adaptation from population-level LLMs to user-specific models.
Further research has explored training LoRAs based on representative users and integrating them into ensembles for target users, enhancing both time and space efficiency~\cite{tan-etal-2024-personalized}.

However, due to the difficulty in collecting user data, data scarcity remains a significant issue. 
For instance, in the LaMP benchmark~\cite{salemi-etal-2024-lamp}, the average number of task-adaptation data tokens exceeds 1,000k, while the average number of tokens per user is only 48k, resulting in extreme data sparsity. 
Additionally, user data distribution follows the Pareto principle~\cite{backhaus1980pareto}, with the top 10\% of users contributing 85\% of the data, further exacerbating this sparsity.
This sparsity makes it difficult for fine-tuning-based methods to learn complex user behavior patterns effectively.

Social science research suggests that the \textbf{meso level}, bridging macro-level (population-level) and micro-level (user-level) analysis, is crucial for understanding the interplay between these two levels~\cite{mcconnell2002meeting,fine2012group,faist2021crucial}. 
Inspired by this, as shown in Figure~\ref{fig:intro}, we propose using group-level LLMs as an intermediary (meso-level) layer between macro- and micro-level LLMs. 
Users with similar preferences and backgrounds can be grouped together~\cite{wood1989theory}, allowing group-level LLMs to capture common patterns from a richer dataset. 
Data-sparse users can then benefit from group-level knowledge, enhancing their personalized models. 

In this paper, we propose PROPER (PROgressive PERsonalization), a novel personalized LLM framework that incorporates a group-level LLM and gradually adapts to users via progressive learning~\cite{fayek2020progressive}. 
Our framework is also inspired by \textbf{residual learning}, where group-level preferences are modeled as a residual shift from population-level preferences, while user-level preferences are further captured by individual residuals beyond the group-level model.
Rather than learning user preferences directly from individual user data, we treat personalization as a hierarchical refinement process.
In this framework, the base model learned from population-level adaptation remains fixed, while subsequent group-level adaptation captures only the residual preferences that the population-level model fails to encode, and similarly for user-level adaptation. 
PROPER thus decomposes LLM personalization into three stages: population-level adaptation, group-level adaptation, and user-level adaptation. 
All adaptation stages employ LoRA to improve computational efficiency.
To construct the group-level LLM, we employ a Mixture-of-Experts (MoE) structure \cite{dou-etal-2024-loramoe}, where each expert represents a user group, and users are automatically assigned to appropriate experts by a user-aware router. 
During user-level adaptation, we further introduce a LoRA-aware router that integrates group-level and user-level LoRAs by selecting the most relevant group-level LoRA based on user-level LoRA knowledge.
Experimental results on the LaMP benchmark show that PROPER significantly outperforms all prior fine-tuning-based baselines.

In conclusion, our contributions are three-fold:
\begin{itemize}
    \item \textbf{New Framework}: We are the first to propose a personalized LLM method that introduces a group-level LLM between population-level and user-level LLMs and integrates it into a progressive learning framework. 
    \item \textbf{Group-level LLM Construction}: we enable automatic user grouping via LoRAMoE and user-aware routers, while effectively integrating user and group-level knowledge through a LoRA-aware router.
    \item \textbf{Empirical Performance}: Our method achieves state-of-the-art results, outperforming all existing fine-tuning-based baselines across all tasks in the LaMP benchmark.
\end{itemize}

\section{Method}

\begin{figure*}[th]
    \centering
    \includegraphics[width=\textwidth]{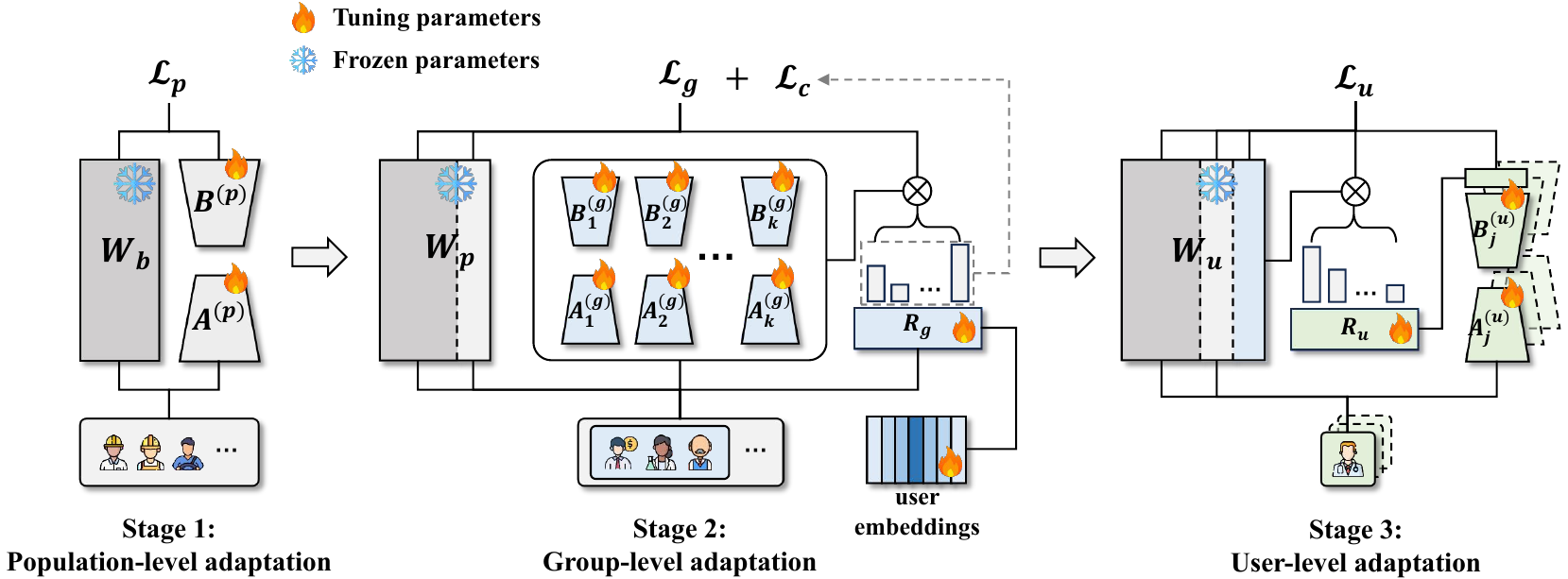}
    \caption{Overview of the training process of PROPER, which consists of three steps: \textbf{(1) Population-level adaptation}, where task information is learned via regular LoRA training; \textbf{(2) Group-level adaptation}, where group-level preferences are learned by LoRAMoE; \textbf{(3) User-level adaptation}, where user preference is learned into user-specific LoRA. The LoRAs are applied to the FFN layers while other components of the Transformer blocks are omitted for simplicity.}
    \label{fig:method}
\end{figure*}

\paragraph{Task Formulation}
Following previous studies~\cite{tan-etal-2024-democratizing, tan-etal-2024-personalized}, we define the task objective of personalized LLM as generating a user-specific response $r_u$ for a given user $u$ at time $t$, based on the user query $q_u$ and the user history $H_u$: $r_u = \text{LLM}(q_u|H_u)$, where $H_u=\{h_u\}$ and each history entry can take one of two forms: either a task-specific query-response pair $h_u=(q_u,r_u)$ or plain text $h_u=t_u$. 
For fine-tuning-based personalized LLMs, the goal is to capture user preferences through user-specific parameter $\Theta_u$, refining the model's response generation as follows:
\begin{equation}
    r_u = \text{LLM}(q_u|\Theta_u).
\end{equation}

\paragraph{Overall Framework}
As shown in Figure~\ref{fig:method}, PROPER is based on progressive learning that consists of three stages: 
\begin{itemize}
    \item Stage 1 (Population-Level Adaptation): The model learns population-level preference for specific tasks using standard LoRA.
    \item Stage 2 (Group-Level Adaptation): The population-level LoRA from Stage 1 is kept fixed, while group-level preferences are learned using a combination of LoRA and MoE.
    \item Stage 3 (User-Level Adaptation): The LoRAs from the previous two stages remain unchanged, and user-specific LoRA is trained to capture individual user preferences.
\end{itemize}

\subsection{Population-Level Adaptation}
Though LLMs are trained on large-scale data, they are not inherently optimized for personalized tasks. 
Following the methods of LaMP~\cite{salemi-etal-2024-lamp}, we first fine-tune the backbone LLM using task-specific query-answer pairs to align it with population-level task preference.

To improve the computational efficiency, we employ Low-rank Adaptation (LoRA)~\cite{hu2022lora} across all LLM adaptation stages.
LoRA assumes that the weight updates during fine-tuning have a low intrinsic rank, allowing them to be decomposed into two smaller matrices. 
Formally, the update process of the feed-forward network (FFN) block in a Transformer can be expressed as:
\begin{equation}
    o = Wx = W_b + \Delta Wx,
\end{equation}
where $o$ denotes the output hidden states, $x$ denotes the input hidden states, $W_b$ is the parameters of the backbone LLM, $\Delta W$ denotes the updated parameter during training.
LoRA approximates $\Delta W \in \mathbb{R}^{d_{in} \times d_{out}}$ using two low ranked matrices $A \in \mathbb{R}^{r \times d_{out}}$ and $B \in \mathbb{R}^{d_{in} \times r}$: $\Delta W \approx BA$, where the rank $r$ is much smaller than $d_{in}$ and $d_{out}$. 

Thus, in the population-level adaptation stage, parameter updates are formulated as:
\begin{equation}
    o = W_b x + \frac{\alpha}{r} B^{(p)} A^{(p)} x,
\end{equation}
where $\Omega_p = \{B^{(p)},A^{(p)}\}$ denotes the population-level LoRA parameters, $r$ is the rank of LoRA components.
To control the learning rate of LoRA components, $\alpha$ is introduced as a scaling factor, which is applied consistently across all adaptation stages.

The population-level LoRA is trained via fine-tuning using the cross-entropy loss:
\begin{equation}
\mathcal{L}_p = \sum_i\text{CE}\{\text{LLM}(q_i|\Omega_p),r_i\}, 
\end{equation}
where $r_i$ is ground-truth response, and $\hat{r}_i=\text{LLM}(q_i|\Omega_p)$ is model-generated output. 
The optimal LoRA parameters are obtained by:
\begin{equation}
\hat{B}^{(p)}, \hat{A}^{(p)} = \arg\min_{\Omega_p} \mathcal{L}_p.
\end{equation}
Finally, the learned parameters $(\hat{B}^{(p)}, \hat{A}^{(p)})$ are merged into the backbone parameters for the next training stage. 
\begin{equation}
    W_p = W_b + \hat{B}^{(p)} \hat{A}^{(p)},
\end{equation}
where $W_p$ is the updated weights of the population-level LLM.

\subsection{Group-Level Adaptation}
Group-level adaptation aims to group users based on shared preferences and learn distinct parameters for each group. 
To achieve this, we employ LoRAMoE~\cite{dou-etal-2024-loramoe}, where each group is represented by an expert, which can be represented as:
\begin{equation}
    o = W_p x + \sum^k_{i=1} \omega_i B_i^{(g)} A_i^{(g)} x,
\end{equation}
where $\{B_i^{(g)} A_i^{(g)}\}_{i=1}^k$ are the group-level LoRA parameters, $k$ is the number of experts, and $\omega_i$ is the weight assigned to the $i$-th expert by the routing mechanism. 
Since manually defining user groups is impractical, we propose to assign users to groups dynamically through a routing mechanism. 
Specifically, we introduce a user-aware router for group-level adaptation by merging the regular router that takes the text embedding $x$ as input with another router that takes the user embedding $u$ as input:
\begin{equation}
\begin{split}
    \omega(x) &= \text{softmax}(h), \\
    h &= \text{softmax}(x M_g) + \text{softmax}(u M_u), 
\end{split}
\end{equation}
where $u \in \mathbb{R}^d$ represents user embeddings that are randomly initialized and updated during training, and $M_g$ and $M_u$ are learnable weight matrices for text router and user router respectively.

MoE training often suffers from unbalanced expert weights, where the model overly relies on a few active experts while neglecting the others. 
To mitigate this, an auxiliary loss is typically applied to balance expert selection~\cite{dou-etal-2024-loramoe, luo2024moelora, liu2024when}. 
In our case, however, enforcing uniform expert selection would lead to redundant group preferences, reducing the effectiveness of user-group differentiation. 
Inspired by P-tailor~\cite{dan2024p}, we introduce a constraint loss that encourages the router to assign distinct expert weights to different users.
Suppose the router weight for user $u_i$ with input $x$ is $\omega_{u_i}(x) = [\omega_1,...,\omega_k]$, then the constraint loss is defined as:
\begin{equation}
\begin{split}
   s_{(i,j)} &=  \omega_{u_i}^T\omega_{u_j}, \\
   \mathcal{L}_c &= \sum_{i \neq j}|s_{(i,j)}|,
\end{split}
\end{equation}
where $s_{(i,j)}$ measures the cosine similarity between the router weights of $u_i$ and $u_j$, encouraging diversity in group assignments.

Following Stage 1, the group-level adaptation parameters, $\Omega_g=\{B_i^{(g)} A_i^{(g)}\}_{i=1}^k \cup \{u\} \cup \{W_g,W_u\}$, are updated and merged into the existing parameters: 
\begin{equation}
\begin{split}
    \mathcal{L}_g &= \sum_i\text{CE}\{\text{LLM}(q_i|\Omega_g),r_i\}, \\
    \hat{B}_j^{(g)}, \hat{A}_j^{(g)} &= \arg\min_{\Omega_g} \mathcal{L}_g, \\
    W_g &= W_p + \frac{a}{r} \sum^k_{j=1} \omega_j B_j^{(g)} A_j^{(g)},
\end{split}
\end{equation}
where $\hat{B}_j^{(g)}, \hat{A}_j^{(g)}$ are the optimized group-level LoRA parameters. 
Here we adopt the idea of residual learning, where group-level preferences can be regarded as a shift from population-level preferences.
Thus, $W_p$ (from population-level adaptation) remains fixed, ensuring that group-level adaptation only captures residual preferences that the population-level adaptation did not model.

\subsection{User-Level Adaptation}
With population- and group-level preferences learned, user-specific adaptations are now regarded as fine-grained modifications to these broader preferences, learned from limited personal data. 
In this stage, following~\cite{tan-etal-2024-democratizing}, we assign a unique LoRA to each user. 
\begin{equation}
    o = W_u x + B_j^{(u)} A_j^{(u)} x,
\end{equation}
where $\{B_j^{(u)}, A_j^{(u)}\}$ are user-specific LoRA for user $j$. 

While the user-aware router in Stage 2 captures user embeddings, its primary function is to guide the group-level experts for user allocation.
That is, the router in Stage 2 is not directly optimized for individual users.  
To address this, we propose a new LoRA-aware router that dynamically integrates group-level LoRAs and user-level LoRAs. 
\begin{equation}
\begin{split}
    \beta_u(x) &= \text{softmax}(W_lh_u), \\
    h_u &= \text{LoRA}_u(x),
\end{split}
\end{equation}
where $\text{LoRA}_u$ represents the learned user-specific LoRA, $h_u$ is the hidden state that passes $x$ though $\text{LoRA}_u$. 
With such implementation, the LoRA-aware router captures both the user-specific LoRA information and the input information.
The final parameters for user-level adaptation for user $u_j$ are trained as follows: 
\begin{equation}
\begin{split}
    \mathcal{L}^{(j)}_p &= \sum_i \text{CE}\{\text{LLM}(q^{(j)}_i|\Omega^{(j)}_p),r^{(j)}_i\}, \\
    \hat{B}_j^{(u)}, \hat{A}_j^{(u)} &= \arg\min_{\Omega^{(j)}_p} \mathcal{L}^{(j)}_p, \\
    W^{(j)}_u &= W_g+B_j^{(u)} A_j^{(u)}+ \sum^k_{m=1} \beta_m B_m^{(g)} A_m^{(g)},
\end{split}
\end{equation}
where $\{q^{(j)}_i,r^{(j)}_i\}$ are user-spefic data for user $u_j$, and $W^{(j)}_u$ is the merged user-specific parameters for $u_j$. 
Combined with other parameters in the Transformer blocks, the final personalized LLM for user $j$ is:
\begin{equation}
    r_u = \text{LLM}(q_u|\Theta_{u_j}).
\end{equation}

\begin{table*}[th]
\centering
\small
\resizebox{2\columnwidth}{!}{
\begin{tabular}{llllllllll}
\toprule
\multicolumn{1}{c}{\multirow{3}{*}{\textbf{Task}}} & \multicolumn{1}{c}{\multirow{3}{*}{\textbf{Metric}}} & \multicolumn{3}{c}{\textit{Prompt-based}}& \multicolumn{5}{c}{\textit{Fine-tuning-based}}       \\
\cmidrule(lr){3-5}\cmidrule(lr){6-10}
\multicolumn{1}{c}{}   & \multicolumn{1}{c}{}     & \multicolumn{1}{c}{\multirow{2}{*}{ICL}} & \multicolumn{1}{c}{\multirow{2}{*}{RAG}} & \multicolumn{1}{c}{\multirow{2}{*}{PAG}} & \multicolumn{2}{c}{\textbf{OPPU}}      & \multicolumn{3}{c}{\textbf{PERPRO}}     \\
\multicolumn{1}{c}{}   & \multicolumn{1}{c}{}     & \multicolumn{1}{c}{}  & \multicolumn{1}{c}{}  & \multicolumn{1}{c}{}  & \multicolumn{1}{c}{kv} & \multicolumn{1}{c}{mlp} & \multicolumn{1}{c}{Stage 1} & \multicolumn{1}{c}{Stage 2} & \multicolumn{1}{c}{Stage 3} \\ \midrule
LAMP-1: PERSONALIZED   & Acc ↑ & .650  & .659  & .756  & .683& .658 & .674     & .663     & \textbf{.691}\\
CITATION IDENTIFICATION& F1 ↑  & .647  & .657  & .755  & .682& .651 & .669     & .667     & \textbf{.687} \\ \midrule
LAMP-2M: PERSONALIZED  & Acc ↑ & .499  & .587  & .534  & .600& .613 & .593     & .701 & \textbf{.747}     \\
MOVIE TAGGING   & F1 ↑  & .441 & .512 & .476   & .493  & .528 & .552     & .611     & \textbf{.666}     \\ \midrule
LAMP-3: PERSONALIZED   & MAE ↓ & .259  & .214  & .321  & .179& .223    & .250     & .196     & \textbf{.178}\\
PRODUCT RATING & RMSE ↓& .590  & .535  & .582  & .443& .490    & .517     & .500     & \textbf{.422}\\ \midrule
LAMP-4: PERSONALIZED   & R-1 ↑ & .187  & .191  & .187  & .191& .197 & .193     & .197     & \textbf{.214}\\
NEWS HEADLINE GEN.     & R-L ↑ & .168  & .172  & .168  & .171& .179 & .174     & .180     & \textbf{.192}\\ \midrule
LAMP-5: PERSONALIZED   & R-1 ↑ & .478  & .505  & .486  & \textbf{.519} & .464 & .491     & .490 & .488\\
SCHOLARLY TITLE GEN.   & R-L ↑ & .418  & .445  & .429  & .442& .419 & .438   & .440 & \textbf{.445}\\ \midrule
LAMP-7: PERSONALIZED   & R-1 ↑ & .524  & .568  & .542  & .539& .513 & .528     & .533     & \textbf{.543}     \\
TWEET PARAPHRASING     & R-L ↑ & .474  & .521  & .501  & .483 & .467 & .481     & .487     & \textbf{.504}     \\ \bottomrule
\end{tabular}
}
\caption{The comparison results of PROPER against baselines on LaMP benchmark. ↑ indicates the higher values are better, ↓ indicates the lower values are better. The best results under \textit{fine-tuning-based} setting are in \textbf{Bold}.}
\label{tab:main}
\end{table*}

\section{Experimental Setup}

\paragraph{Datasets} 
Following the previous work~\cite{tan-etal-2024-democratizing,tan-etal-2024-personalized,zhuang2024hydra}, we conduct experiments using the Large Language Model Personalization (LaMP) benchmark~\cite{salemi-etal-2024-lamp}.
LaMP evaluates LLM personalization across seven tasks, including three classification tasks (personalized citation identification, movie tagging, and producing rating) and four generation tasks (personalized news headline generation, scholarly title generation, Email subject generation, and tweet paraphrasing)\footnote{We exclude the LaMP-6: Email subject generation task due to restricted access to private data.}.
To make a fair comparison with OPPU~\cite{tan-etal-2024-democratizing}, we adopt the same 100 test users selected by OPPU in user-level adaptation, while all other users for population-level and group-level adaptation.
Additional task details can be found in Appendix~\ref{sec:appendix-task}.

\paragraph{Baselines} 
We compare PROPER with both non-personalized and personalized baselines including the prompt-based methods and fine-tuning-based methods. 
For the backbone LLM used in PROPER and all baselines, we employ Llama-2-7B to make a fair comparison with prior work. 
Further details on the baseline can be found in Appendix~\ref{sec:appendix-baseline}.
Implementation details and hyperparameters settings can be found in Appendix~\ref{sec:appendix-implementaion} and Appendix~\ref{sec:appendix-hyperparameters}.

\paragraph{Evaluation Metrics}
In line with LaMP~\cite{salemi-etal-2024-lamp}, we use accuracy and F1-score for classification tasks (LaMP-1 and LaMP-2M), Mean Absolute Error (MAE) and Root Mean Squared Error (RMSE) for LaMP-3, and adopt ROUGE-1 and ROUGE-L for text generation tasks (LaMP-4, LaMP-5, LaMP-7). Higher values indicate better performance for all metrics, except for MAE and RMSE (where lower values are better).

\section{Experimental Results}
In this section, we present comprehensive experiments conducted on LaMP. 
Through an in-depth analysis of the results, we aim to address the following Research Questions (\textbf{RQs}):
\begin{itemize}
    \item \textbf{RQ1}: How does PROPER perform compared to baseline models in a standard setting?
    \item \textbf{RQ2}: How effectively does PROPER handle data sparsity?
    \item \textbf{RQ3}: What impact do different architectural structures and components have on model performance?
    \item \textbf{RQ4}: What is the trade-off between personalization quality and computational cost?
    \item \textbf{RQ5}: How effectively do group experts capture group-level preferences?
    \item \textbf{RQ6}: How does PROPER perform in qualitative evaluations?
\end{itemize}

\subsection{Main Results}

To answer \textbf{RQ1}, we compare the performance of PROPER with other baseline models in the regular setting. 
The results, shown in Table~\ref{tab:main}, demonstrate that PROPER consistently outperforms the baseline methods, highlighting its strong capability in personalization. 
We observe the following:

\paragraph{PROPER delivers universal improvements.} 
Compared to OPPU, PROPER achieves significant improvements across all six tasks, with minimal additional computation and storage overhead. 
Specifically, for the LaMP 2M: Personalized Movie Tagging task, PROPER yields relative improvements of 24.5\% in accuracy and 35.1\% in F1-score compared to OPPU. 
For the other tasks, PROPER also demonstrates consistent improvements, with an average relative improvement of 5.47\%. 
We do not include PER-PCS~\cite{tan-etal-2024-personalized} for comparison, as it focuses primarily on time and space efficiency and ties with OPPU in performance.

\paragraph{Progressive learning results in consistent improvements across stages.}
To investigate the improvements at each stage, we present the detailed performance from Stage 1 (population-level adaptation) to Stage 3 (full model). 
The results show consistent improvements from one stage to the next, demonstrating the effectiveness of progressive training. 
From Stage 1 to Stage 2, the average relative improvement is 4.69\%, and from Stage 2 to Stage 3, the average relative improvement is 5.02\%. These results highlight the importance of both stages.

\paragraph{Performance depends on the task and user history format.}
As shown in the results, the performance of prompt-based and fine-tuning-based methods varies depending on the task.
For classification tasks (LaMP-1, LaMP-2M, and LaMP-3), fine-tuning-based methods generally perform better, with PROPER outperforming the prompt-based baselines by significant margins in LaMP-2M and LaMP-3.
For generation tasks, fine-tuning-based and prompt-based methods perform similarly, with PROPER being outperformed by the RAG baseline in LaMP-7 and the PAG baseline in LaMP-1.
We hypothesize that for classification tasks, LLMs struggle to learn the mapping between input texts and output labels using limited examples through in-context learning. 
In contrast, fine-tuning-based methods can learn the mapping more effectively through supervised learning. For generation tasks, LLMs can easily learn the style and background from the examples and tailor the input query to user preferences.

\subsection{Low-Resource Results}

To answer the \textbf{RQ2}, we compare the performance of PROPER with other baseline models in an extremely low-resource setting, where we select the top 100 \textbf{inactive} users for personalization. 
The results are shown in Table~\ref{tab:low-resource}. It can be observed that PROPER outperforms OPPU on all metrics, demonstrating the effectiveness of the model in data sparsity.
The results are consistent with the main results where PROPER outperforms other baselines by the largest margin on LaMP-2M, which is the most data-sparse task in the LaMP benchmark with an average user history length of 55.6.
\begin{table}[t]
\centering
\resizebox{0.7\columnwidth}{!}{
\begin{tabular}{lcccc}
\hline
\multicolumn{1}{l}{\multirow{2}{*}{\textbf{Settings}}} & \multicolumn{2}{c}{\textbf{LaMP-3}} & \multicolumn{2}{c}{\textbf{LaMP-5}} \\
\multicolumn{1}{c}{}  & MAE ↓         & RMSE ↓          & R-1 ↑           & R-L ↑  \\ \hline
OPPU                  & .327          & .644            & .472            & .439   \\
\rowcolor{mycell} PROPER                & \textbf{.303} & \textbf{.582}   & \textbf{.522}   & \textbf{.483}  \\ \hline
\end{tabular}}
\caption{Performances comparison between PROPER and OPPU under an extremely low-resource setting on LaMP-3 and LaMP-5. The best results are in \textbf{Bold}.}
\label{tab:low-resource}
\end{table}

\subsection{Ablation Studies}

To answer \textbf{RQ3}, we evaluate PROPER under different settings on LaMP 2M (personalized movie tagging) and LaMP 7 (personalized tweet paraphrasing). 
For Stage 2 (group-level adaptation), we examine the effectiveness of the user-aware router by replacing it with a regular router based solely on the input state $x$, and assess the impact of removing the constraint loss.
Note that we do not compare these versions under full-stage training (including Stage 3), as PROPER uses progressive learning. 
In this framework, earlier trained stages are fixed, so any underperformance in Stage 2 would likely carry over into Stage 3.
For Stage 3 (user-level adaptation), we evaluate the effectiveness of the LoRA-aware router by removing it and investigate the impact of replacing progressive training with end-to-end training. 
In the end-to-end setup, we jointly train the group-level experts with the user-specific LoRAs, using the user-aware router and constraint loss.
As shown in Table~\ref{tab:ablation}, removing or replacing components in PROPER leads to a significant drop in performance on both tasks, demonstrating the effectiveness of the designed components.

\begin{table}[t]
\centering
\resizebox{0.95\columnwidth}{!}{
\begin{tabular}{lcccc}
\toprule
\multicolumn{1}{l}{\multirow{2}{*}{\textbf{Settings}}} & \multicolumn{2}{c}{\textbf{LaMP-2M}} & \multicolumn{2}{c}{\textbf{LaMP-7}}  \\\cmidrule(lr){2-3}\cmidrule(lr){4-5}
\multicolumn{1}{c}{}  & Acc↑   & F1↑    & R-1↑   & R-L↑   \\ \midrule
\rowcolor{mycell} Stage 2 (group-level adaptation)  & \textbf{.701}& \textbf{.611}& \textbf{.527}& \textbf{.481}\\
\quad w/ regular router    & .659 & .564 & .513 & .515 \\
\quad w/o constraint loss  & .686 & .602  & .564  & .472 \\ \hline
\rowcolor{mycell} Stage 3 (user-level adaptation)   & \textbf{.747}& \textbf{.666}& \textbf{.542}& \textbf{.504}\\
\quad w/o LoRA-aware router& .726  & .645  & .534  & .483  \\
\quad End-to-end training  & .723 & .644  &  .528 & .477 \\ \bottomrule
\end{tabular}
}
\caption{Ablation Studies for PROPER on LaMP 2M and LaMP 7 tasks. 
The best results in the corresponding stage are in \textbf{Bold}.
}
\label{tab:ablation}
\end{table}

\begin{figure*}[t]
    \centering
    \includegraphics[width=\textwidth]{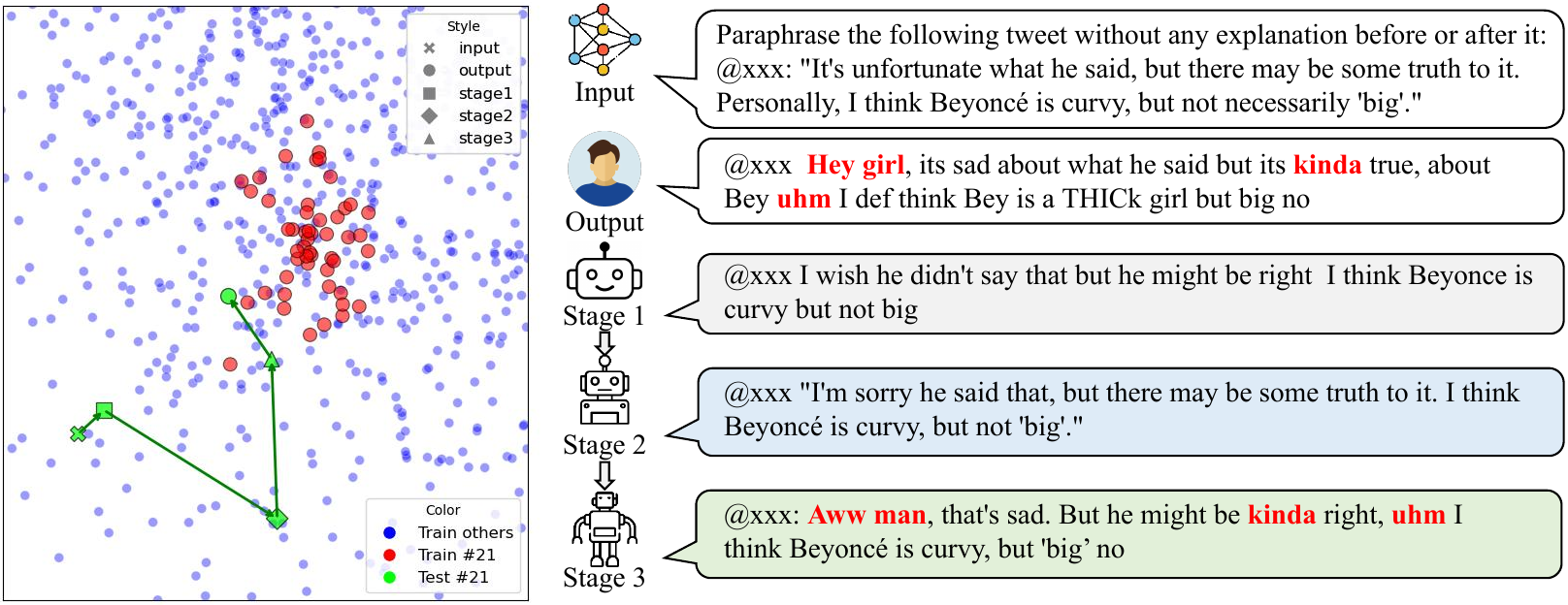}
    \caption{The case study on LaMP-7: Personalized Tweet Paraphrasing task. The figure on the left shows the visualization of text embeddings for user \#21. 
    The \textcolor{green}{green} legends represent the test example and the model output,  \textcolor{red}{\large $\bullet$} represent the training examples for user \#21, while  \textcolor{blue!80}{\large $\bullet$} represent the training examples for other users.}
    \label{fig:case}
\end{figure*}

\subsection{Efficency Analysis}

\begin{figure}[t]
    \centering
    \includegraphics[width=0.48\textwidth]{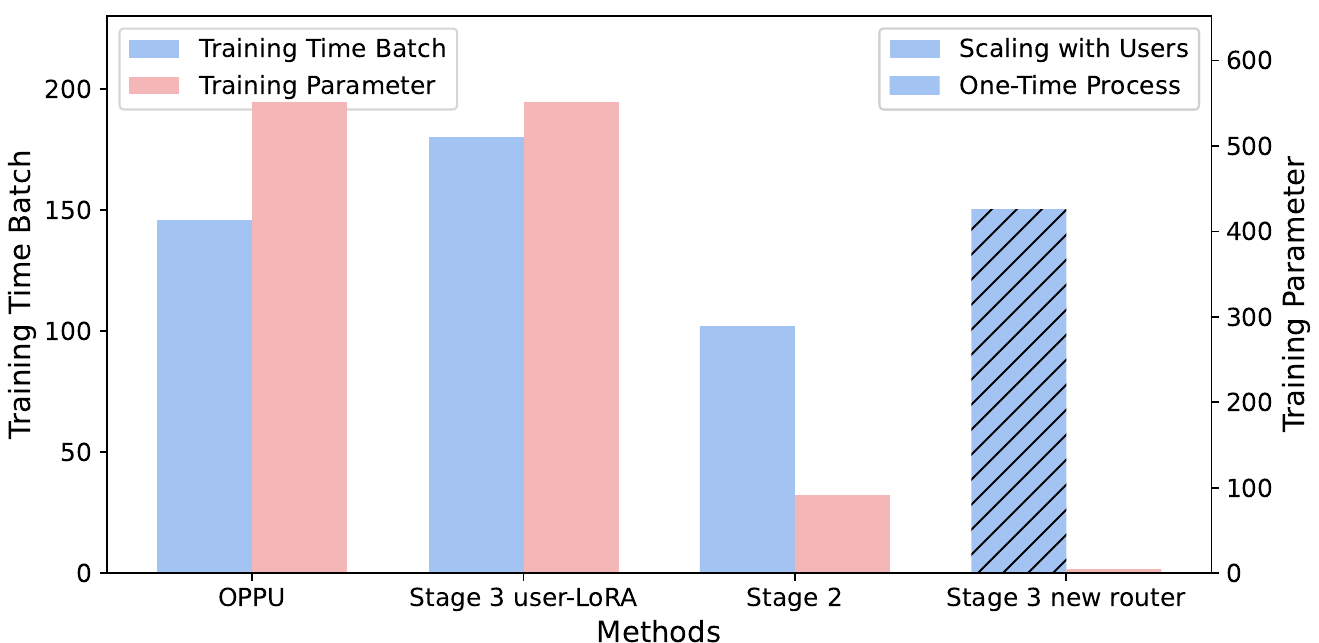}
    \caption{Comparison of training time and training parameters between OPPU and different stages of PROPER, the training time is calculated for 100 test users per batch, and all results are produced with a single NVIDIA A100 GPU (80GB).}
    \label{fig:efficency}
\end{figure}

To answer \textbf{RQ4}, we compare OPPU and different stages of PROPER in terms of trainable parameters and training time. 
As shown in Figure~\ref{fig:efficency}, the main computational and storage load in personalized LLMs is in the user-level adaptation stage, which scales with the number of users. 
Both OPPU and PROPER introduce 552M parameters for 100 users. Regarding training time, PROPER takes slightly longer (180 min per batch) than OPPU (146 min) for 100 users. 
PROPER also introduces two additional components: the group-level adaptation and the LoRA-aware router. 
These components are one-time processes that do not scale with user growth, adding minimal computation (146 min for group-level adaptation and 150 min for LoRA-aware router) and storage overhead (91M and 4M, respectively). 
Despite these additions, PROPER remains efficient overall due to its improvements.

\begin{figure}[t]
    \centering
    \includegraphics[width=0.48\textwidth]{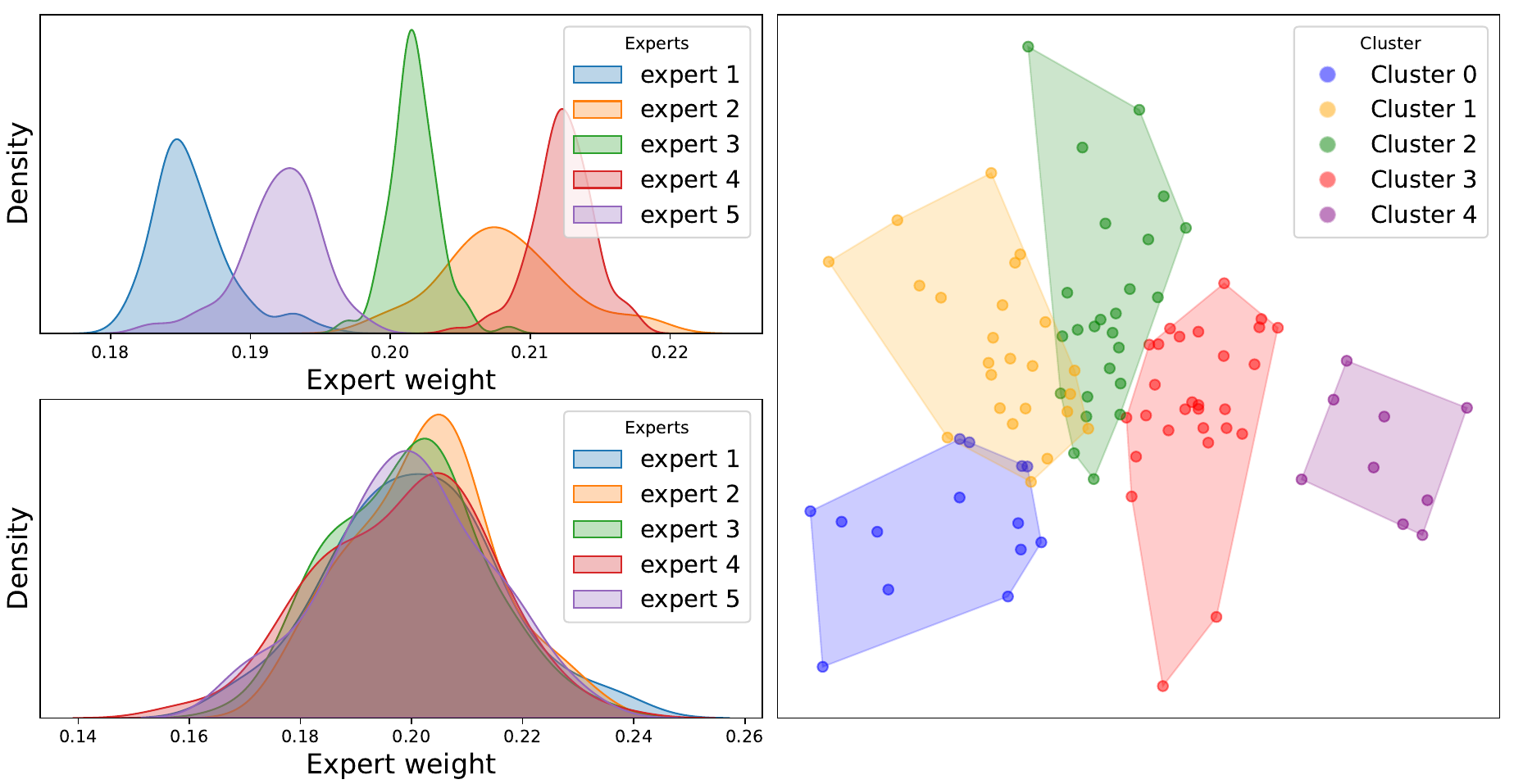}
    \caption{The Visualization of expert weights and user embeddings learned in the group-level adaptation. \textbf{The upper left}: density plot of expert weights with the user-aware router and constraint loss; \textbf{The bottom left}: density plot of expert weights with regular LoRAMoE; \textbf{The right}: Scatter plot of user embeddings after detention reduction, colored by the clusters.}
    \label{fig:visual}
\end{figure}

\subsection{Visualization}

To answer \textbf{RQ5}, we visualize the user embeddings learned in Stage 2 and the expert weights for the group experts.
For the expert weights, we average the weights for each user and compare the density plots of expert weights learned with the user-aware router and constraint loss versus those learned with regular LoRAMoE.
For the user embeddings, we average the embeddings across layers for each user and apply t-SNE~\cite{van2008visualizing} to map them into a 2D space. We then cluster the users into 5 groups based on their averaged expert weights and color the user embeddings according to their cluster.
As shown in Figure~\ref{fig:visual}, the density plot of expert weights with the user-aware router and constraint loss shows five distinct peaks with minimal overlap, indicating that the experts learn distinct group preferences. In contrast, the expert weights from regular LoRAMoE are highly overlapping, suggesting that the experts are learning redundant information.
In the user embedding visualization, we observe that the clustering of user embeddings aligns with the expert weight patterns, demonstrating a clear correlation between the behavior of user embeddings and expert weight distribution.

\subsection{Case Study}

To answer \textbf{RQ6}, we conduct a case study on the LaMP-7: Personalized Tweet Paraphrasing task for user \#21 (user\_id: 13002361) in the test set. 
To demonstrate the effectiveness of progressive learning, we visualize the training and test samples for user \#21, as well as a subset of training samples for other users. 
For text visualization, we use a BERT encoder~\cite{reimers-gurevych-2019-sentence} to generate text embeddings and apply T-SNE for dimensionality reduction.
By comparing the embeddings (left side of Figure~\ref{fig:case}) with the corresponding texts (right side), we observe that the input tweet to be paraphrased has a formal tone, while the user’s tweet (\textit{i.e.}, the target output) is more casual with many non-standard expressions. 
During the population-level adaptation stage, the model's output (Stage 1 output in Figure~\ref{fig:case}) retains a formal tone, and its embedding \textcolor{green!70}{$\blacksquare$} stays close to the input tweet (green cross). 
However, as progressive learning advances, the output becomes more casual, incorporating expressions like "\textcolor{red}{\textbf{kinda}}" (Stage 3 output) and the embeddings \textcolor{green!70}{$\blacktriangle$} move closer to the target output \textcolor{green!70}{$\bullet$} . 
By Stage 3, the model output closely align with user \#21's historical data 
 and other relevant training samples, illustrating the effectiveness of progressive learning in personalization.

\section{Related Work}

\paragraph{Personalized LLMs}
Personalized LLMs can be broadly categorized into two types: prompting-based and fine-tuning-based.
Prompting-based methods augment the LLM’s input prompt with user history while keeping the LLM itself unchanged wiht in context learning~\cite{dai2023uncovering,kang2023llms}. 
Following the idea of Retrieval-Augmented Generation (RAG), subsequent approaches refine this paradigm by retrieving relevant user history for each query~\cite{salemi-etal-2024-lamp,mysore-etal-2024-pearl}. 
Another variant summarizes a user profile from user history and then augments the user prompt with the inferred profile~\cite{richardson2023integrating}.
Fine-tuning-based methods inject user information directly into the LLM’s parameters via fine-tuning. 
\citet{tan-etal-2024-democratizing} introduced OPPU, which assigns each user a specific LoRA module for personalization, while PER-PCS~\cite{tan-etal-2024-personalized} improves efficiency by assembling user-specific LoRA from relevant pieces trained on representative users.
Beyond adapting the LLM itself, \citet{zhuang2024hydra} proposed to fine-tune both rerankers and adapters within a retrieval-based framework to align with black-box LLMs.

\paragraph{Mixture of Experts}
The Mixture-of-Experts (MoE) replaces feed-forward layers with sparsely activated experts, enabling dynamic expert selection per input, which expands capacity without significantly increasing computational cost~\cite{jacobs1991adaptive}.
Previously, the token-level MoE architectures are widely used in pre-trained language models and vision models~\cite{shazeer2017,lepikhin2021gshard,riquelme2021scaling,du2022glam}.
Currently, with the fast development of LLMs, the need for efficient tuning of a model has become more and more important, therefore, many works try to combine MoE with PEFT methods such as LoRA~\cite{hu2022lora}.
The most straightforward way is to combine LoRA and MoE for multi-task learning~\cite{dou-etal-2024-loramoe,luo2024moelora,liu2024when}.
P-Tailor~\cite{dan2024p} proposed a MoE-based role-playing LLM that models the Big Five Personality Traits using specialized LoRA experts and adapts personality traits across topics.

\section{Conclusion}
\label{sec:conclusion}
In this paper, we present PROPER, a novel progressive learning framework for personalized LLMs that addresses the challenge of data sparsity by introducing a group-level adaptation process. By leveraging a Mixture-of-Experts structure and LoRA-based routers, PROPER enables efficient adaptation through population-level, group-level, and user-level stages, bridging the gap between broad and personalized models. Our extensive experiments demonstrate that PROPER significantly outperforms existing state-of-the-art approaches, offering a promising solution for more efficient and effective LLM personalization in diverse applications. Future work will focus on further optimizing group-level adaptations and exploring additional techniques to enhance model scalability and generalizability.

\section*{Limitations}
We identify three key limitations in PROPER. 

First, due to dataset constraints, PROPER evaluates LLM personalization on separate tasks. 
In real-world applications, it would be more practical and beneficial to consider LLM personalization within a multi-task learning framework, where user preferences learned from one task can enhance performance in other tasks. 
Despite this, PROPER can be adapted to multi-task learning, as its LoRAMoE module is inherently suited for such integration. 

Second, PROPER assumes user preferences are static, but in reality, user preferences may evolve over time. 
Future research could focus on dynamically modeling these preferences or developing a framework capable of continually learning from streaming data.

Third, the three-stage process of PROPER introduces additional training parameters and longer training times. 
Although these extra parameters and training processes are manageable and a one-time cost, future work should aim to improve the training and inference efficiency of the progressive learning-based framework.

\section*{Ethical Impact}
Personalized large language models (LLMs), such as the PROPER framework, rely on user-specific data, raising privacy concerns regarding the potential inadvertent disclosure of sensitive information. Strong privacy safeguards, including data anonymization and encryption, must be implemented to protect personal data. Additionally, biases in user data can lead to unfair or prejudiced model outputs, emphasizing the need for diverse, balanced data and debiasing techniques to ensure fairness. Transparency in decision-making processes is essential, allowing users to understand how their data influences personalized outputs and ensuring accountability. Accessibility is another concern, as the computational demands of advanced LLMs may limit adoption among smaller entities and researchers, exacerbating the digital divide. To address this, efforts to make personalized LLMs more accessible, such as resource-efficient models, are crucial. Finally, user autonomy should be respected by allowing individuals to control their data and the level of personalization, ensuring ethical use and avoiding over-dependence on AI-generated content. Addressing these ethical considerations will promote the responsible development and deployment of personalized LLMs, prioritizing privacy, fairness, and accessibility while mitigating potential risks.

\section*{Acknowledgments}
This work was supported by the UK Engineering and Physical Sciences Research Council (EPSRC) through a Turing AI Fellowship (grant no. EP/V020579/1, EP/V020579/2). 
The work was also supported by the National Natural Science Foundation of China (62176053).


\appendix

\section{Appendix}
\label{sec:appendix}

\subsection{Task Details}
\label{sec:appendix-task}
LaMP~\cite{salemi-etal-2024-lamp} provided seven separate tasks to benchmark LLM personalization, following~\cite{tan-etal-2024-democratizing,tan-etal-2024-personalized}, we describe the task details as follows to help readers gain a better understanding of the task format.
\begin{itemize}
    \item \textbf{LaMP-1: Personalized Citation Identification}: is a binary text classification task. The input query $x$ is a paper title written by user $u$, along with two candidate paper titles, the output $y$ is the number of the candidate paper that $u$ will cite in $x$. The user history contains titles and abstracts of the publications of user $u$.
    
    \item \textbf{LaMP-2M: Personalized Movie Tagging}: is a 15-way text classification task. The labels are pre-defined movie types. The input query $x$ is the movie description, and the output $y$ is the tag that user $u$ will give based on $x$. The user history contains the user’s historical movie-tag pairs $(x,y)$.
    
    \item \textbf{LaMP-3: Personalized Product Rating}: is a 5-way text classification task. The input query $x$ is the review text written by user $u$, and the output $y$ is the corresponding score that user $u$ will given based on $x$. The user history is the previous rating pairs $(x,y)$ of user $u$.

    \item \textbf{LaMP-4: Personalized News Headline Generation}: is a text generation task to test the model’s ability to capture the stylistic patterns in personal data. The input query $x$ is the content of a news from the author $u$, and the output $y$ is the news headline generated by user $u$. The user history is the historical article-title pairs $(x,y)$ from author $u$.
  
    \item \textbf{LaMP-5:Personalized Scholarly Title Generation}: similar to LaMP-4, it is a text generation task to test personalized text generation tasks in different domains. The input query $x$ is the abstract of a paper, and the output $y$ is the title generated by user $u$. The user history is the historical abstract-title pairs $(x,y)$ from author $u$.
    
    \item \textbf{LaMP-7:Personalized Tweet Paraphrasing} is also a text generation task that tests the model’s capabilities in capturing the stylistic patterns of authors. The input query $x$ is a normalized tweet, and the output $y$ is the original tweet from user $u$. The user history is the historical tweets from author $u$.
\end{itemize}

\subsection{Baseline Details}
\label{sec:appendix-baseline}
We present the task details as follows to help readers gain a better understanding of the task format.
\begin{itemize}
    \item ICL (In-Context Learning)~\cite{dai2023uncovering}: This method randomly selects user historical records to augment the input query for LLM. In this paper, we take the results reported from ~\cite{tan-etal-2024-democratizing}.
    \item RAG (Retrieval Augmentation Generation)~\cite{salemi-etal-2024-lamp}: Following the retrieval-augmented personalization method presented in LaMP, the user’s query is augmented with top k retrieved items from the corresponding user’s history corpus. In the paper, we take the results that the number of retrieval items k=1 from ~\cite{tan-etal-2024-democratizing}.
    \item PAG (Profile Augmentation Generation)~\cite{richardson2023integrating}: In the PAG-based method, the user’s input sequence would concatenate the user’s profile summarizing the user’s preference and behavior patterns. In the implementation by ~\cite{tan-etal-2024-democratizing}, The vicuna-7B model is employed for user profile generation and the model is further enhanced with the retrieval augmentation. In the paper, we take the results that the number of retrieval items k=1 from ~\cite{tan-etal-2024-democratizing}.
    \item OPPU (kv)~\citet{tan-etal-2024-democratizing}: The original implementation of OPPU, where the LoRA components are placed on the KV-Cahce in the transformer blocks. We do not include the hybrid integration of the prompt-based method and fine-tuning-based method posted in \cite{tan-etal-2024-democratizing} because the integration of the prompt-based method violates the principle of privacy of the fine-tuning-based method. 
    \item OPPU (mlp): We implement another version of OPPU by changing the placement of LoRA components from the KV-Cahce in the transformer blocks to the mlp projection layers. 
\end{itemize}

\subsection{Implementaion Details}
\label{sec:appendix-implementaion}
% We maintain the original weights
% of the backbone architecture unchanged and integrated Feed-Forward Network(FFN)
% components of all layers.
Following~\cite{tan-etal-2024-personalized}, we
incorporate trainable low-rank adapters into the $W_q$, $W_k$, $W_v$, $W_o$ , setting the rank $r$=8.
Additionally, we set the factor of $\alpha$ to 16, using a learning rate of 3e-4 and adapt batch size of 2 at stage 1 population adaptation for all tasks.

For the subsequent stages,
We maintain the original weights
of the backbone which merged LoRA from stage 1 unchanged and integrate low-rank adapters~\cite{hu2022lora,zhang-etal-2024-star} into the Feed-Forward Network(FFN) components of all layers.
Specifically, in the LLaMA2 model~\cite{touvron2023llama}, the FFN layer utilizes the SwiGLU structure~\cite{shazeer2020glu}, which consists of three components: down projection, up projection, and gate.

The number of experts is set to be 5 for all tasks based on the primary experiments. 
These low-rank adapters are configured with a rank of 4 and a factor of $\alpha$
set to 8, alongside a dropout rate of 0.05 to mitigate overfitting.
The model parameters are optimized
by AdamW~\cite{loshchilov2018decoupled}. 
We use a batch size of 1 to facilitate the identification of specific users and a learning rate of 2e-4 for all tasks.
Our implementation leverages the \texttt{PyTorch}\footnote{\url{https://github.com/pytorch/pytorch}} framework, \texttt{HuggingFace Transformers}\footnote{\url{https://github.com/huggingface/transformers}}~\cite{wolf-etal-2020-transformers} and \texttt{PEFT}\footnote{\url{https://github.com/huggingface/peft}} library.
All experiments are carried out with an NVIDIA A100 80GB GPU.
% Training 100 personal PEFT sequentially took around 12 minutes to 12 hours depending on the size of the behavior history corpus and the sequence length per history item.
% LoRA details, and hyperparameter details (such as learning rate).

\subsection{HyperParameter Analysis}
\label{sec:appendix-hyperparameters}

As the number of experts $k$ serves as a very important hyperparameter in PROPER, we perform an analysis on the number of experts 2, 5, and 8 on the LaMP-2M task.
As shown in Figure~\ref{fig:efficency}, with an increase of experts number, the performance increase then decrease, suggesting a peak around 5. 
This can be explained that with fewer experts, the group-level pattern can not be fully distinguished and some experts learn the overlapped pattern, while with many experts, the experts are redundant and the group-level patterns are overfitted.
As it is costly to perform a fine-grained hyperparameter search for all tasks, we set the number of experts as 5 for all the tasks.

\begin{figure}[t]
    \centering
    \includegraphics[width=0.48\textwidth]{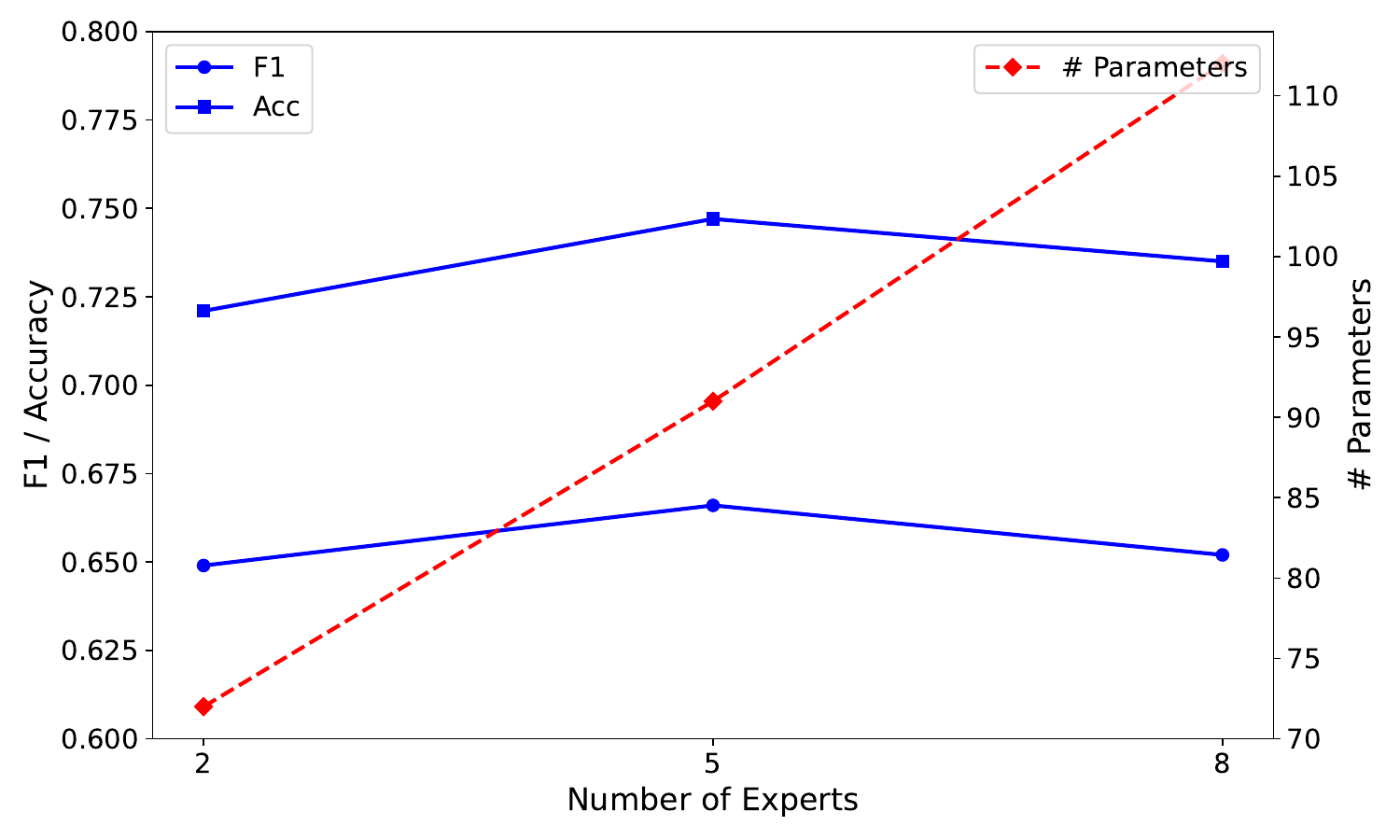}
    \caption{Compartion of the different number of experts on LaMP-2M.}
    \label{fig:efficency}
\end{figure}

\end{document}